\documentclass[journal,twoside,web]{ieeecolor}
\usepackage{generic}
\usepackage{cite}
\usepackage{amsmath,amssymb,amsfonts}
\usepackage[table,xcdraw]{xcolor}
\usepackage{graphicx}
\usepackage{hyperref}
\usepackage{textcomp}
\usepackage{bbm}
\usepackage{booktabs}
\usepackage{multirow}
\usepackage{pythonhighlight}
\usepackage{makecell}
\usepackage{soul}
\usepackage{algorithm,algpseudocode}
\def\BibTeX{{\rm B\kern-.05em{\sc i\kern-.025em b}\kern-.08em
    T\kern-.1667em\lower.7ex\hbox{E}\kern-.125emX}}
\begin{document}
\title{CrossMatch: Enhance Semi-Supervised Medical Image Segmentation with Perturbation Strategies and Knowledge Distillation}
\author{Bin Zhao, Chunshi Wang, and Shuxue Ding
\thanks{This work is supported in part by the Youth Science Foundation of Guangxi Natural Science Foundation (Grant No.2023GXNSFBA026018), the Project of Improving the Basic Scientific Research Ability of Young and Middle-Aged Teachers in Universities of Guangxi Province (Grant No.2023KY0223),  the  National Natural Science Foundation of China (Grant No.62076077),   and the Guangxi Science and Technology Major Project (Grant No.AA22068057). (Corresponding author: Shuxue Ding.) }
\thanks{Bin Zhao, Chunshi Wang and Shuxue Ding are with School of Artificial Intelligence, Guangxi Colleges and Universities Key Laboratory of AI Algorithm Engineering, Guilin University of Electronic Technology, Guilin 541004, Guangxi, China(e-mail:zhaobinnku@mail.nankai.edu.cn, 2101630316@mails.guet.edu.cn,  sding@guet.edu.cn).}
\thanks{Bin Zhao and Chunshi Wang contribute equally to this paper.}}

\maketitle

\begin{abstract}
Semi-supervised learning for medical image segmentation presents a unique challenge of efficiently using limited labeled data while leveraging abundant unlabeled data. Despite advancements, existing methods often do not fully exploit the potential of the unlabeled data for enhancing model robustness and accuracy. In this paper, we introduce CrossMatch, a novel framework that integrates knowledge distillation with dual perturbation strategies, image-level and feature-level, to improve the model's learning from both labeled and unlabeled data. CrossMatch employs multiple encoders and decoders to generate diverse data streams, which undergo self-knowledge distillation to enhance the consistency and reliability of predictions across varied perturbations. Our method significantly surpasses other state-of-the-art techniques in standard benchmarks by effectively minimizing the gap between training on labeled and unlabeled data and improving edge accuracy and generalization in medical image segmentation. The efficacy of CrossMatch is demonstrated through extensive experimental validations, showing remarkable performance improvements without increasing computational costs. Code for this implementation is made available at \url{https://github.com/AiEson/CrossMatch.git}.   
\end{abstract}

\begin{IEEEkeywords}
Semi-supervised segmentation; Self-knowledge distillation; Image perturbation
\end{IEEEkeywords}

\section{Introduction}\label{sec:introduction}
\IEEEPARstart{S}{emantic} segmentation, as a precise classification technique at the pixel level, plays a vital role in the field of medical image analysis. Especially when dealing with complex three-dimensional CT and MRI data, although fully supervised learning methods can achieve high-precision segmentation results, their applications are severely limited by the high cost of manual annotation and the complexity of operation. In order to overcome this bottleneck, semi-supervised medical image segmentation methods have emerged, and {demonstrated} great potential \cite{han2024deep}. The core of these approaches lies in the effective combination of a small amount of annotated data and a large amount of unlabeled data, aiming to reduce the high cost of annotation and achieve accurate segmentation while promoting widespread application in clinical and other scenarios.

The main challenge in semi-supervised learning (SSL) is how {to exploit the potential of unlabeled data effectively.} Recent research has shifted from relying on adversarial training mechanisms based on Generative Adversarial Networks (GANs)~\cite{gan2017,gan2019} to incorporating various methods, including consistency regularization and self-training~\cite{cps,c_regular_nips_2021,fixmatch,st++,unimatch}. In particular, collaborative teaching and mutual learning paradigms~\cite{CAML,dml,mcnet,mcnet_plus,dmd} have proven to be highly promising strategies, often involving the parallel training of two models. Knowledge distillation strategies have also been widely employed to optimize model structures, enabling efficient training and good performance by simplifying models.

In handling unlabeled image data, the application of both image-level and feature-level perturbations has become a common strategy. Image-level perturbations, such as random rotations, scaling, flipping and color adjustments, enhance model robustness to input variations through controlled deformations and modifications of the input images. Moreover, more complex image-level perturbations like CutMix~\cite{yun2019cutmix} and MixUp~\cite{zhang2017mixup} create new training samples by blending regions between images and combining them at the pixel level, thus simulating a more diverse data distribution and further improving the model's generalization to unseen data.
Feature-level perturbations, particularly those applied to features extracted by the Encoder, have not been fully explored and hold substantial potential. This approach introduces weak to strong feature perturbations during the Decoder decoding process, utilizing the model's prediction consistency under various perturbation conditions to train the model, which ensures stability in performance when the model faces the same image segmentation tasks. For example, feature-level perturbations can be achieved by adding random noise, applying various types of Dropout, etc.~\cite{unimatch,liu2022perturbed}. These perturbations not only simulate potential variations in the data but also promote generalization in the model's deep feature abstraction and decoding process, thereby achieving more accurate and robust predictive performance on unlabeled data.

Knowledge Distillation (KD) has demonstrated significant potential in semi-supervised learning for medical image segmentation~\cite{kd_hiton,dmd}. Typically, KD involves a pre-trained teacher model and a student model that needs to be learned. However, Self-KD methods~\cite{self_kd,beyourowntch} primarily rely on soft labels generated within a single model to guide the training process instead of depending on traditional hard labels or an additional teacher model. These methods use the model's self-generated predictions during training as guidance, refining the model's feature extraction and classification capabilities through iterative processes. This self-teaching method not only reduces dependence on costly manually annotated data but also significantly enhances the model's adaptability and prediction accuracy on unlabeled data. Self-KD promotes deeper feature learning and more stable model behaviour by reinforcing the model's reliance on its own predictions. Particularly for medical imaging data, this strategy effectively improves model robustness and accuracy when dealing with highly variable and individually distinct medical images.

Inspired by Self-KD and image perturbation, we have designed an innovative self-training consistency regularization framework called CrossMatch for semi-supervised medical image segmentation. This framework employs a range of image-level and feature-level perturbations from weak to strong and explores the potential of unlabeled data through a more systematic and in-depth approach. Specifically, CrossMatch applies two types of image-level and two types of feature-level perturbations to unlabeled data to create four distinct data streams. 
These data streams vary in accuracy of output prediction depending on the degree of perturbation to which they are subjected, where the stronger streams guide the weaker ones. In this process, image-level perturbations are implemented as applications of different encoders, while feature-level perturbations are used to generate varied outputs for the same decoder. Through these perturbations, CrossMatch engages in internal knowledge distillation by leveraging the model's consistency across different perturbation intensities, which not only optimizes the model's learning from unlabeled data but also enhances its generalization capability.
CrossMatch ensures the stability and accuracy of model outputs, thereby exhibiting superior performance in applications requiring high precision, such as medical image segmentation.

In summary, our contributions are fourfold:
\begin{itemize}
  \item[(1)] We propose a consistency regularization framework based on knowledge distillation and image perturbations, which focuses on the exploration of unlabeled data and the transfer of self-knowledge.
  \item[(2)] We equate different feed-forward flows to different encoders and decoders, applying the concept of knowledge distillation to semi-supervised semantic segmentation.
  \item[(3)] We compute adjacent Self-KD losses between the same decoders, which can bridge the capability gap between the teacher and student models.
  \item[(4)] Experimental results on four benchmark datasets demonstrate that CrossMatch achieves significant performance improvements compared to previous state-of-the-art methods.
    
\end{itemize}

\section{Related Work}
\subsection{Semi-Supervised Learning}
In the field of SSL, a key challenge is designing effective supervision signals for unlabeled data. Currently, there are two main strategies to address this issue: entropy minimization~\cite{em1,em2,em3,em4} and consistency regularization~\cite{fixmatch,remixmatch,cr1,unimatch}.
{Entropy minimization, a concept utilized in semi-supervised learning settings, involves measuring the difference in entropy between the outputs of teacher and student models. By minimizing this difference, we ensure that the student model's predictions are confident and closely aligned with those of the teacher model. This strategy aids in stabilizing the learning process by fostering more reliable and consistent predictions from the student model.
Incorporating this into the broader context of entropy minimization and consistency regularization, we can see how these strategies synergize. Entropy minimization is appreciated for its straightforward approach of automatically assigning pseudo-labels to unlabeled data. These pseudo-labels are then used to retrain the model alongside labeled data, enhancing the model's ability to generalize from limited labeled information. Consistency regularization complements this by ensuring that a model's predictions for the same unlabeled sample remain consistent across different perturbations. This consistency is crucial, as it helps in reducing overfitting and improving the robustness of the model under varying conditions.}
For example, FixMatch \cite{fixmatch} combines the advantages of entropy minimization and consistency regularization to apply strong perturbations to unlabeled images and use the predictions from their weakly perturbed versions to guide model training. Advanced methods like FreeMatch~\cite{wang2022freematch} further refine this strategy, providing rigorous mathematical justification for its motivation and using thresholds to filter out low-confidence labels, thereby enhancing the model's accuracy and reliability.

Our CrossMatch draws on the basic framework of FixMatch without any bells and whistles. It only uses the most common way to verify the theoretical effectiveness of this method and also demonstrates its important value in practical applications.

\subsection{Semi-Supervised Semantic Segmentation}
Semi-supervised learning based methods have achieved exciting results in classification tasks, of which several works have been further developed for semantic segmentation. A popular class of methods~\cite{CAML,uamt,copy_paste} is based on the Mean Teacher~\cite{mean_teacher} setting. For instance, UA-MT~\cite{uamt} introduces a self-aware model of uncertainty to design thresholds to filter out uncertain regions between teachers and students to get more meaningful and reliable predictions. BCP~\cite{copy_paste} notes that in semi-supervised learning, the distributions learned in labeled and unlabeled data are not consistent and proposes a symmetric approach to use both kinds of data so as to maintain the consistency between the two distributions, thus allowing the model to learn common features. CAML~\cite{CAML} pays further attention to the potential of labeled data and proposes a Correlation Aware Mutual Learning framework to utilize labeled data to guide the extraction of information from unlabeled data. CPS~\cite{cps} utilizes a cross-teacher module to simultaneously reduce the coupling among peer networks and the error accumulation between teacher and student networks.

Another mainstream class of semi-supervised segmentation methods is based on the idea of co-training. The networks learn together and transfer knowledge to each other~\cite{dml, dmd}.  To transfer knowledge efficiently between networks, knowledge distillation is also a common strategy in semi-supervised semantic segmentation~\cite{dmd}. Besides, some method uses pseudo segmentation maps obtained from one network to supervise the other one~\cite{ctt}. MC-Net~\cite{mcnet} and MC-Net+~\cite{mcnet_plus} use a shared encoder for feature extraction and then feed the features into multiple decoders with the same structure but different parameters to get multiple outputs. All these methods require multiple networks, encoders or decoders for training.

Methods based on self-training have begun to evolve rapidly since FixMatch~\cite{fixmatch}  introduced consistency regularization to self-training, and FixMatch has gradually become the baseline for many methods. DTC~\cite{dtc} uses a dual-task deep network to jointly predict pixel segmentation maps and geometrically-aware level-set representations of a target by introducing dual-task consistency regularization between level-set derived segmentation maps and directly predicted segmentation maps for labeled and unlabeled data. SASSNet~\cite{sassnet} introduces a multi-task deep network that jointly predicts semantic segmentation and symbolic distance maps (SDM) of object surfaces while introducing adversarial loss in order to capture shape-aware features. URPC~\cite{urpc} enhances pyramid-consistent regularization using multi-scale uncertainty correction for more efficient semi-supervised medical image segmentation. SS-Net~\cite{ssnet} addresses the challenges of semi-supervised medical image segmentation by simultaneously exploring pixel-level smoothness and inter-class separation. UniMatch~\cite{unimatch} achieves better segmentation results by consistency regularization using multiple strongly augmented branches and a dual-stream perturbation feature perturbation. Our CrossMatch also follows this single-stage framework, i.e., there is only one model in our approach. Unlike the above works, our CrossMatch introduces Self-KD and feature perturbation into semi-supervised medical image segmentation, achieving efficient self-knowledge transfer under a broader perturbation space.

\section{Method}

\subsection{Preliminaries}
Semi-supervised medical image segmentation aims to fully explore an unlabeled image set $\mathcal{D}^u = \{x^u_1, \ldots, x^u_n\}$ and integrate it with a labeled image set $\mathcal{D}^l = \{(x^l_1, y^l_1), \ldots, (x^l_n, y^l_n)\}$ that contains limited annotations for precise semantic segmentation. The performance of series methods like FixMatch~\cite{fixmatch} largely depends on well-designed image-level perturbation strategies. Specifically, each unlabeled input is subjected to two types of perturbations: $\mathcal{A}^w$ denotes a weak perturbation operator, and $\mathcal{A}^s$ denotes a strong perturbation operator. Given an unlabeled input $x^u$, we have
\begin{equation}
\left\{
\begin{array}{l}
    x^{w}=\mathcal{A}^{w}\left(x^{u}\right)  \\
     x^{s}=\mathcal{A}^{s}\left(\mathcal{A}^{w}\left(x^{u}\right)\right),
\end{array}
\right.
\end{equation}
where  $x^{w}$ and $x^{s}$ represent the weakly perturbed image and the strongly perturbed image, respectively.

\begin{figure}
    \centering
    \includegraphics[width=\linewidth]{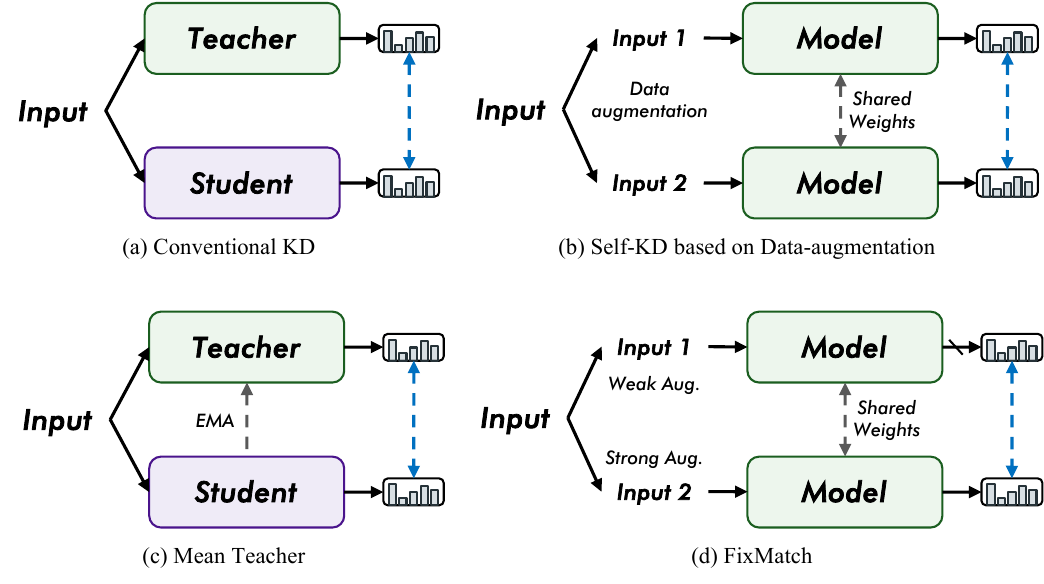}
    \caption{Comparison of different types of KD and SSL methods. (a) Traditional KD  requires pre-training of the teacher model. (b) Self-KD based on data augmentation. (c) Mean Teacher. (d) FixMatch.}
    \label{fig:kd_and_ssl}
\end{figure}

\subsection{Knowledge Distillation}
In machine learning tasks, Kullback-Leibler (KL) divergence is often used to measure the discrepancy between different probability distributions. In knowledge distillation, it is commonly employed to gauge the performance gap between teacher and student models,
\begin{equation}
\mathcal L_{kd}^{KL}\left(p^{w_i}, p^{w_j}\right)=K L\left(\sigma(p^{w_i}/T), \sigma(p^{w_j}/T)\right),
\end{equation}
where $p^{w_i}$ and $p^{w_j}$ represent the probability distribution outputs by the teacher and student models for unlabeled samples, respectively. Here, $\sigma(\cdot)$ denotes the softmax function, which transforms logits into normalized probability distributions, and $T$ is a hyperparameter known as the temperature coefficient. 
{This is a scaling factor used in the softmax function during knowledge distillation to control the smoothness of output probabilities. A higher temperature results in softer probability distributions, facilitating the transfer of richer information between the teacher and student models.}
Thus, within the framework of knowledge distillation, the KL divergence loss function $\mathcal L_{kd}^{KL}$ aims to minimize the difference between the probability distributions of the teacher model $w_i$ and the student model $w_j$ that after softmax processing and temperature reduction. 
{In this way, the student model can emulate the teacher model's 'soft' output predictions, thereby facilitating the effective transfer of complex and high-quality knowledge.}

DMD~\cite{dmd} delves into knowledge distillation methods specifically for semi-supervised medical image segmentation and proposes to use  Dice loss as an alternative to KL divergence loss. This approach effectively addresses the common issue of foreground and background class imbalances in segmentation tasks.  Compared to KL divergence loss, Dice loss can more aptly handle such imbalances, thereby enhancing the model's segmentation performance,
\begin{equation}
\mathcal L_{kd}^{Dice}\left(p^{w_i}, p^{w_j}\right)=\text{Dice}\left(\sigma(p^{w_i}/T), \sigma(p^{w_j}/T)\right).
\end{equation}

As illustrated in Figure \ref{fig:kd_and_ssl}, a careful comparison of KD methods and SSL methods reveals remarkable similarities in the structure, design and development of the networks. Based on this observation, we hypothesize that KD methods can be readily adapted to SSL tasks.

\subsection{Feature Pertubation}
The performance of FixMatch~\cite{fixmatch} and its related works, such as UniMatch~\cite{unimatch} and ReMixMatch~\cite{remixmatch}, largely depends on the effectiveness of the well-designed image-level perturbation strategies. As mentioned earlier, pseudo-labels generated from the weakly perturbed images $x^{w}$ are used to supervise the strongly perturbed images $x^{s}$ to achieve consistency  learning. The greater the difference in the degree of perturbation between $x^{w}$ and $x^{s}$, the larger the perturbation space during training. Generally, the perturbation space should be within an appropriate range according to~\cite{yuan2021simple}: too small a difference may diminish the effect of consistency regularization, while excessive perturbation can have a catastrophic impact on the clean data distribution.

Although image-level perturbations have {been} widely used in numerous methods, their performance in semi-supervised image segmentation tasks highly depends on how researchers meticulously tailor perturbation schemes for specific datasets to ensure an appropriate perturbation space is constructed. This process often involves a high demand for expert knowledge and trial-and-error costs, especially in the field of medical image processing, where finding suitable perturbation strategies can become one of the main challenges demanding significant effort.

To mitigate the issues mentioned above, the literature \cite{unimatch,liu2022perturbed} suggests perturbing the high-dimensional features of $x^w$ at the bottleneck section of the segmentation network by using different levels of perturbation to create varied feed-forward flows. Segmentation models typically employ an encoder-decoder structure, where $e$ denotes the encoder and $d$ denotes the decoder. For FixMatch, the weak perturbation feed-forward flow for an unlabeled sample $x^u$ can be represented as:
\begin{equation}
    x^u \to \mathcal{A}^w \to e \to d \to p^{w},
\end{equation}
where $x^u \to \mathcal{A}^w = x^w$. Based on this format, we can consider inserting a new perturbation $\mathcal{P}^r$ between $e \to d$ to achieve a larger perturbation space and obtain a  new perturbation feed-forward flow:
\begin{equation}\label{eq:weak_stream}
    x^u \to \mathcal{A}^w \to e \to \mathcal{P}^r \to d \to p^{w}_r,
\end{equation}
where $r$ differentiates the intensity of feature perturbations, which will be detailed later. Similarly, the feature perturbation flow for the strongly perturbed input can be represented as:
\begin{equation}\label{eq:strong_stream}
    x^u \to \mathcal{A}^w \to \mathcal{A}^s \to e \to \mathcal{P}^r \to d \to p^{s}_r.
\end{equation}

For consistent notation, the aforementioned flows can be succinctly expressed as follows:
\begin{equation}
\left\{
\begin{array}{l}
    x^w \to e \to \mathcal{P}^r \to d \to p^{w}_r
\\
x^s \to e \to \mathcal{P}^r \to d \to p^{s}_r.
\end{array}
\right.
\end{equation}
Let $\mathcal{P}^{n}$ denote no feature perturbation applied, then $p^w$ can be obtained through the flow $x^w \to e \to \mathcal{P}^{n} \to d \to p^{w}$. Based on this, we can compute the loss function for FixMatch with feature perturbation as:
\begin{equation}
    \frac{1}{B_{u}} \sum \mathbbm{1}\left(\max \left(p^{w}\right) \geq \tau\right)\left(\mathrm{H}\left(p^{w}, p^{s}_r\right)+\mathrm{H}\left(p^{w}, p^w_r\right)\right),
\end{equation}
where $\mathrm{H}(\cdot)$ denotes the entropy minimizing the discrepancy between two probability distributions. $\mathbbm{1}(\cdot > \tau)$ is the indicator function for confidence-based thresholding with the threshold $\tau$. 
{$\max(\cdot)$ represents taking the maximum value along the channel dimension to obtain the confidence map.}
{$B_{u}$ is the batch size for unlabeled data.}

\subsection{Multiple Encoders and Decoders}\label{sec:multi_en_de}
In Eq.~\ref{eq:weak_stream} and Eq.~\ref{eq:strong_stream}, we re-examine  $\mathcal{A}^w \to e$ and $\mathcal{A}^w \to \mathcal{A}^s \to e$, whose aim is to achieve consistency in model predictions by applying varying degrees of image-level perturbations. In contrast, Mean Teacher (MT)~\cite{mean_teacher} and UA-MT~\cite{uamt} achieve a similar effect by the utilization of    Exponential Moving Average (EMA), which allows a model to derive one or more other models for training. By integrating this different method of achieving consistency with knowledge distillation, we can consider different levels of perturbations as encoders with varying capabilities. Hence, let $e^w = \mathcal{A}^w \to e$ represent the weak perturbation encoder and $e^s = \mathcal{A}^w \to \mathcal{A}^s \to e$ represent the strong perturbation encoder, where $e^w$ is clearly outperforms $e^s$, that is, $e^w$ is less perturbed and its resulting prediction is clearly more accurate.

Similarly, based on the relationship between the encoders and perturbations mentioned above, the newly introduced feature perturbations can be viewed as perturbations to the decoder's capabilities, where $d^r = \mathcal{P}^r \to d$ represents the high-dimensional features entering the decoder being perturbed by $\mathcal{P}^r$. Consistent with the form of the encoders, here we propose using both strong and weak feature perturbations, namely $\mathcal{P}^w$ and $\mathcal{P}^s$, thus yielding three different decoders $d^{n}, d^w$ and $d^s$. Consequently, we now have two equivalent encoders and three equivalent decoders.

\subsection{CrossMatch}
Fig.~\ref{fig:kd_and_ssl} demonstrates various KD and SSL methods, revealing a high similarity between them, thus prompting the idea of integrating knowledge distillation into SSL tasks. The purpose of knowledge distillation is to transfer more accurate knowledge to another network model. Self-KD represents a unique distillation mode where the student model learns from knowledge generated from its outputs, typically involving the backpropagation of deep information to guide the training of earlier layers. This approach incorporates image-level perturbations to achieve varying capabilities, as depicted in Fig.~\ref{fig:kd_and_ssl} (b), a process very similar to that in Fig.~\ref{fig:kd_and_ssl} (c).
\begin{figure}
    \centering
    \includegraphics[width=\linewidth]{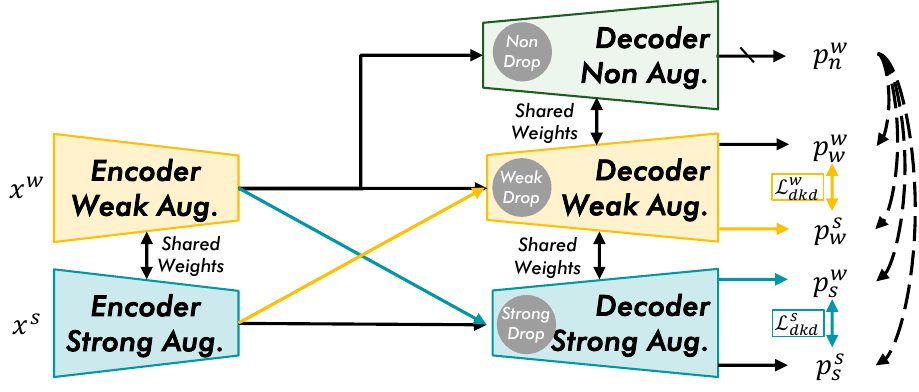}
    \caption{Overview of our proposed CrossMatch. CrossMatch integrates the core ideas of Self-KD and SSL by enhancing performance through the derivation and mutual distillation of multiple encoder-decoder architectures.}
    \label{fig:overview}
\end{figure}

Based on this, we propose CrossMatch, whose overall structure is depicted in Fig.~\ref{fig:overview}. CrossMatch employs multiple different encoders and decoders, namely $e^w$, $ e^s$, $d^{n}$, $ d^w$ and $d^s$ as mentioned in Sec.~\ref{sec:multi_en_de}, to generate diverse outputs. These combinations produce outputs denoted as $p^i_j$. Specifically, $x^u$ passes through the following feed-forward flow to form different outputs:
\begin{equation}
    x^u \to e^i \to  d^j \to p^{i}_j,
\end{equation}
where $e^i \in \{e^w, e^s\}$ and $d^j \in \{d^{n}, d^w, d^s\}$. Notably, $p^w_n$ experiences the least perturbation and is the most accurate. Here, we designate $p^w_n$ as the Teacher, with all other outputs, which have undergone feature perturbations, acting as students. The Teacher is required to impart knowledge to all students, leading to the following teacher distillation loss:
\begin{equation}\label{eq:tkd}
    {\mathcal L_{tkd}} = \frac{1}{B_{u}} \sum \mathbbm{1}\left(\max \left(p^w_n\right) \geq \tau\right)\sum_i \sum_j \mathrm{H}\left(p^w_n, p^i_j\right).
\end{equation}

Observing that a decoder outputs two segmentation results with varying degrees of perturbation, we can facilitate mutual distillation between these outputs. Specifically, we consider $p^w_w$ and $p^w_s$ as teaching assistants, each imparting knowledge to $p^s_w$ and $p^s_s$, respectively. These teaching assistants are relative to the same decoder hence this is referred to as decoder distillation loss:
\begin{equation}\label{eq:deckd}
    {\mathcal L_{dkd}} = \frac{1}{B_{u}} \sum \mathbbm{1}\left(\max \left(p^w_j\right) \geq \tau\right) \sum_j  \mathrm{H}\left(p^w_j, p^s_j\right).
\end{equation}
Eq.~\ref{eq:tkd} and Eq.~\ref{eq:deckd} correspond to the black and colored arrows in Fig.~\ref{fig:overview}, respectively.

In practical implementation, as shown in Fig.~\ref{fig:impl}, we also introduce two image-level strong perturbations ($x^{s_1}$ and $x^{s_2}$), which are applied with the same degree of perturbation but differ in perturbation parameters. This aligns with the principles of contrastive learning and has been proven meaningful for our tasks in previous works~\cite{unimatch,uniperb1,uniperb2}.

Finally, by combining the supervised loss $\mathcal L_{sup}$, the image-level perturbation loss $\mathcal L_{ip}$, we can derive the total loss:
\begin{equation}\label{eq:total_loss}
    \mathcal{L}_{total} = \mathcal{L}_{sup} + \mathcal{L}_{ip} + (1-\eta)\mathcal{L}_{tkd} + \eta\mathcal{L}_{dkd},
\end{equation}
where $\mathcal{L}_{sup}$ consists of Dice and CrossEntropy losses, and $\mathcal{L}_{ip}$ denotes the supervision of $p^w_n$ over $p^{s_1}$ and $p^{s_2}$ as shown in Fig.~\ref{fig:impl}, which involves calculating Dice for the two strongly perturbed unlabeled predictions and averaging them. $\eta$ is used to balance the proportions between the two distillation losses.

\subsection{Performance-Friendly Implementation}\label{sec:efficient_compute}

{CrossMatch's multi-encoder/decoder setup is straightforward, but training costs rise due to multiple forward propagations for perturbations. Therefore, we propose effective and equivalent implementation methods.}

{For image-level perturbations, we apply data augmentations (weak $\mathcal{A}^w$ and strong $\mathcal{A}^s$) on CPUs before the encoder stage,} {leveraging parallel processing to avoid additional computation during model iterations.
As shown in Fig.~\ref{fig:overview}, CrossMatch uses four feature perturbations from two decoders. Let $h^i = x^u \to e^i$ represent intermediate features from different encoders and $h^i_{w,s} \in \mathbbm{R}^{2B\times H\times W\times C}$ for more efficient computation, stacking features in the Batch dimension to preserve the independence of perturbation outcomes.}

\begin{figure}
    \centering
    \includegraphics[width=\linewidth]{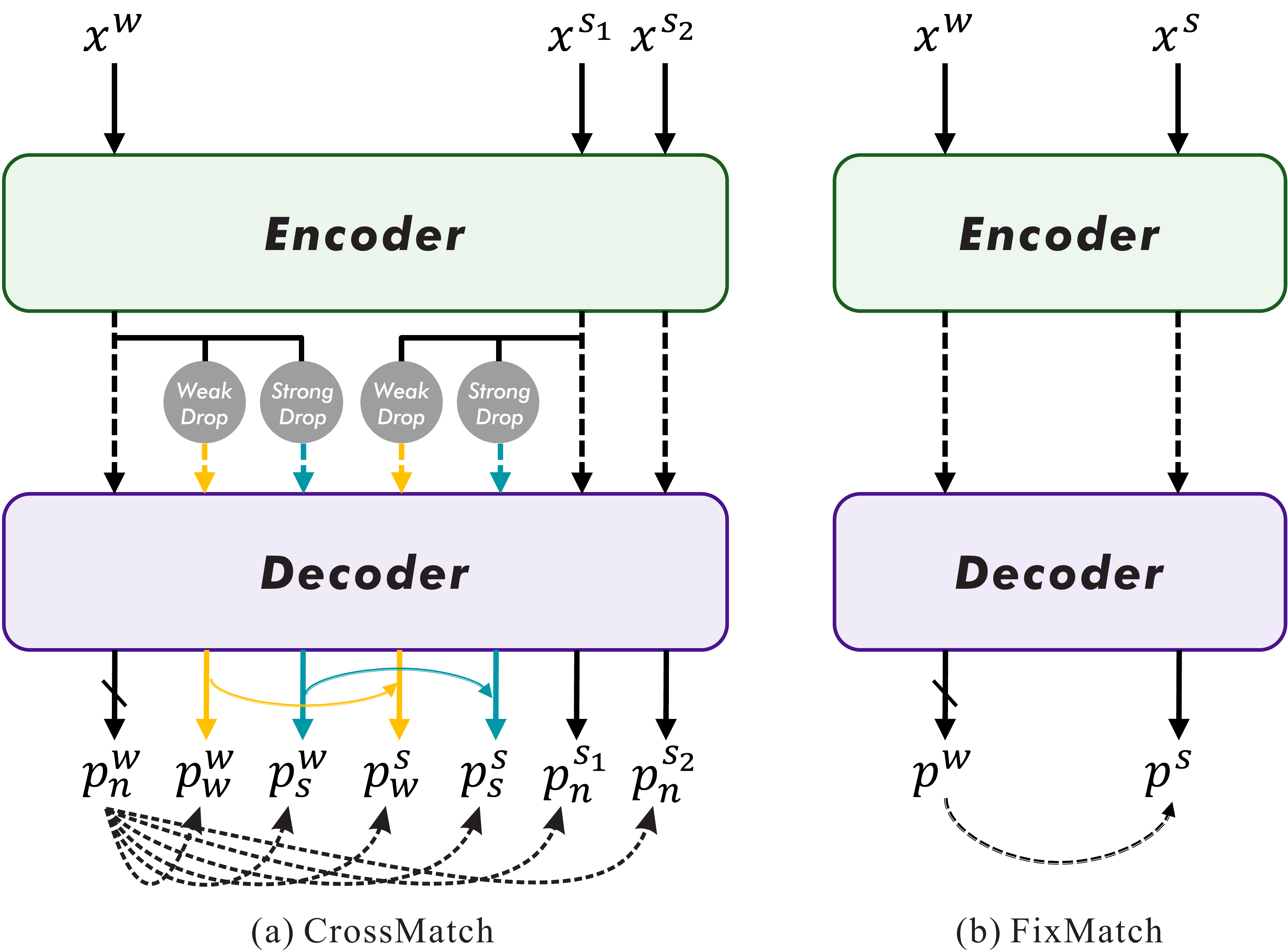}
    \caption{(a) Our proposed CrossMatch method with Weak Drop denoting $\mathcal{P}^w$ and Strong Drop denoting $\mathcal{P}^s$. (b) FixMatch.}
    \label{fig:impl}
\end{figure}

As shown in Fig.~\ref{fig:impl}, our CrossMatch does not introduce any additional parameter overhead and adheres to the principles of Self-Training and Self-KD. It only uses image-level and feature perturbations to expand the perturbation space of FixMatch, proving to be more efficient than the EMA method and introducing knowledge distillation into semi-supervised learning tasks. Theoretically, multiple encoders and decoders are introduced, but the implementation employs a more efficient coding method, achieving significant performance improvement while ensuring computational friendliness.

\subsection{The pseudocode of CrossMatch}
In summary, we present a self-training framework for multiple encoders and decoders based on knowledge distillation and provide a performance-friendly implementation. Algorithm~\ref{alg:alg1} provides the pseudocode of  CrossMatch.

\section{Experiments}
\subsubsection{Dataset}
In this study, we utilize the 2018 Left Atrium Segmentation Challenge (LA\footnote[1]{\url{www.cardiacatlas.org/atriaseg2018-challenge/}}) as a platform to evaluate the proposed CrossMatch. The challenge provides data consisting of 3D MRI scans and their corresponding left atrium segmentation masks, divided into training and validation sets in an $80/20$ ratio, with an isotropic resolution of $0.625 \times 0.625 \times 0.625 ~mm^3$.
Furthermore, we extend our experimental work to the Automatic Cardiac Diagnosis Challenge (ACDC\footnote[2]{\url{www.creatis.insa-lyon.fr/Challenge/acdc/}}). To ensure a fair comparison with previous works, we follow the same experimental setup when reporting the performance on the validation set.

{Additionally, we include the Pancreas-CT\footnote[3]{\url{cancerimagingarchive.net/collection/pancreas-ct/}} dataset, collected by the NIH Clinical Center for Pancreas Segmentation, which includes 82 contrast-enhanced 3D CT scans of the abdomen (\(512 \times 512\) resolution) with slice thicknesses ranging from \(1.5\) to \(2.5~\text{mm}\). For image preprocessing, all axes are resampled to an isotropic resolution of \(1.0~\text{mm}\) and cropped to Hounsfield Unit (HU) values between \([-125, 275]\).}

{In the end, we use the ISIC 2018\footnote[4]{\url{https://challenge.isic-archive.com/data/\#2018}} dataset, which includes 3,694 dermoscopic RGB images. There are 2,594 images in the training set, 100 images in the validation set and 1,000 images in the test set. The original image sizes range from 540 $\times$ 722 to 4499 $\times$ 6748 pixels. We resize all images to 256 $\times$ 256 pixels and randomly shuffled the dataset for standardized processing. Depending on the experiment's requirements, we randomly split the training set into 5\%, 10\%, and 20\% subsets for further model training. These preprocessing steps ensure data consistency and provide standardized inputs.}

{All in all, we conduct experiments on four datasets, covering 2D and 3D segmentation of CT, MRI and RGB images, with a wide range of sample sizes, fully demonstrating the outstanding performance of our method.}



\begin{algorithm}[H]
\caption{{Pseudocode of CrossMatch}}\label{alg:alg1}
\begin{algorithmic}[1]
\State \textbf{Input:} Unlabeled data $\mathcal{U} = \{x_i\}_{i=1}^{N}$, encoder $e$, decoder $d$, perturbation functions $\mathcal P^w$ and $\mathcal P^s$, criterion $\mathrm{H}(\cdot)$, hyperparameter $\eta$.
\State \textbf{Output:} Optimized $e$ and $d$.
\For{each batch $\{x_w, x_{s1}, x_{s2}\}$ in $\text{loader}_u$}
    \State $\{h_w,h_s\} \gets e(\{x_w,x_{s2}\})$ \Comment{Features of $x_w,x_{s2}$}
    
    \State $p^w_n \gets h_w$ \Comment{None drop}
    \For{$i \in \{w, s\}$}
        \For{$j \in \{w, s\}$}
            \State $p^i_j \gets d(\mathcal P^j(h_i))$ \Comment{Apply feature perturbations to \( h_i \) and decode.}
        \EndFor
    \EndFor
    
    \State $\{p_{s1}, p_{s2}\} \gets d(e(\{x_{s1}, x_{s2}\}))$

    \State Compute the $\mathcal L_{tkd}$ as shown in Eq.~\ref{eq:tkd}.
    \State Compute the $\mathcal L_{dkd}$ as shown in Eq.~\ref{eq:deckd}.
    \State Compute the $\mathcal L_{total}$ as shown in Eq.~\ref{eq:total_loss}.

    \State Use the optimizer to update $e$ and $d$ through $\mathcal{L}_{{total}}$.
    
\EndFor
\end{algorithmic}
\end{algorithm}

\subsubsection{Implementation Details}
The CrossMatch is implemented based on PyTorch and uses V-Net and U-Net as the baseline networks for experiments on the (LA, Pancreas-CT) and (ACDC, ISIC 2018) datasets, respectively.
{On the LA dataset, CrossMatch use the AdamW~\cite{adamw} optimizer for 9000 iterations, and on the Pancreas-CT dataset, it needs training for 12000 iterations. On the ACDC and ISIC 2018 datasets, it uses the SGD optimizer for 300 and 200 epochs of optimization respectively.}
Different batch sizes are set for different datasets, with LA and Pancreas-CT at 4, ACDC and ISIC-2018 at 12, ensuring an equal number of labeled and unlabeled samples per batch. 
{For image preprocessing, the LA and Pancreas-CT datasets are randomly cropped to $112 \times 112 \times 80$ and $96 \times 96 \times 96$, respectively. The ACDC dataset is cropped to $256 \times 256$, and ISIC-2018 is adjusted to $256 \times 256$. We set $\eta=0.2$ for the Pancreas-CT dataset and $\eta=0.3$ for the other datasets. For LA and Pancreas-CT, $\tau$ is set to 0.85, and for ACDC and ISIC-2018, $\tau$ is set to 0.95.}
For performance evaluation, the LA and Pancreas-CT datasets use a sliding window strategy to achieve comprehensive segmentation of the cardiac area, while the ACDC dataset is evaluated by merging predicted slices into a 3D image. {For ISIC-2018, it is evaluated directly using 2D metrics.} The evaluation metrics, including Dice, Jaccard, 95\% Hausdorff Distance (95HD) and Average Surface Distance (ASD) are used in this paper. {In all CrossMatch experiments, feature perturbations are set as a standard dropout. The dropout rates for weak and strong perturbations are set at 25\% and 75\%, respectively. The selection of dropout type and discussion on dropout rates are elaborated in Sec.~\ref{sec:discussion}.}

Notably, to ensure the fairness of the experiments, our results are calculated using the final model weights rather than the best weights saved during training for all 3D tasks, which also demonstrates the stability of our method.

\begin{figure}
    \centering
    \includegraphics[width=0.9\linewidth]{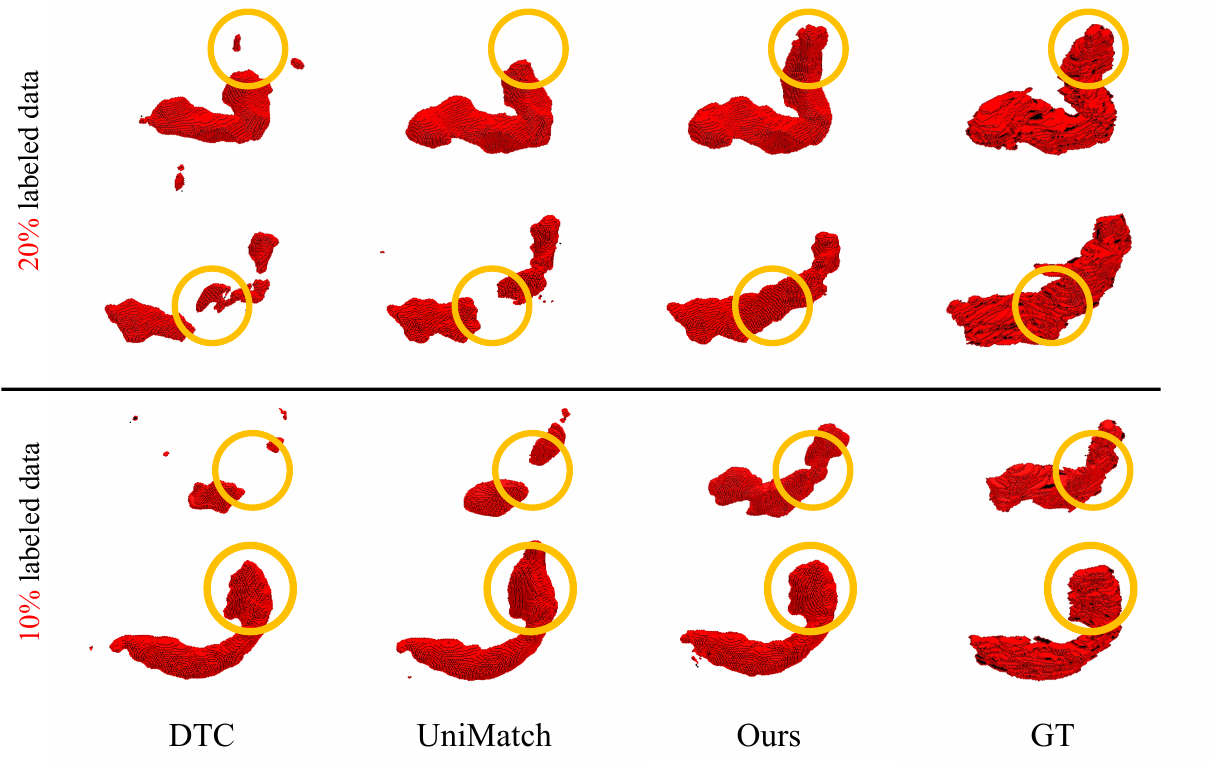}
    \caption{{Visualizations of several semi-supervised segmentation methods with 10\% and 20\% labeled data and ground truth on Pancreas-CT dataset.}}
    \label{fig:pct_3d_result}
\end{figure}

\begin{table*}
\centering\caption{Comparisons on the LA dataset. "$\uparrow$" and "$\downarrow$" indicate the larger and the smaller the better, respectively.}\label{tab:la_results}
\begin{tabular}{@{}ll|ll|l|l|l|l@{}}
\toprule
\multicolumn{2}{c|}{} &
  \multicolumn{2}{c|}{\#Scans used} &
  \multicolumn{4}{c}{{Metrics}} \\ \cmidrule(l){3-8} 
\multicolumn{2}{c|}{\multirow{-2}{*}{{Method}}} &
  {Lab.} &
  \multicolumn{1}{l|}{{Unlab.}} &
  {Dice(\%)$\uparrow$} &
  {Jaccard(\%)$\uparrow$} &
  {95HD(voxel)$\downarrow$} &
  {ASD(voxel)$\downarrow$ }\\ \midrule
V-Net &
   &
  4(5\%) &
  0 &
  43.32 &
  31.43 &
  40.19 &
  12.13 \\
V-Net &
   &
  8(10\%) &
  0 &
  79.99 &
  68.12 &
  21.11 &
  5.48 \\
V-Net &
   &
  16(20\%) &
  0 &
  86.03 &
  76.06 &
  14.26 &
  3.51 \\
V-Net &
   &
  80(All) &
  0 &
  91.14 &
  83.82 &
  5.75 &
  1.52 \\ \midrule
UA-MT~\cite{uamt} &
  (MICCAI'19) &
  4(5\%) &
  76(95\%) &
  78.07 &
  65.03 &
  29.17 &
  8.63 \\
SASSNet~\cite{sassnet} &
  (MICCAI'20) &
   &
   &
  79.61 &
  67.00 &
  25.54 &
  7.20 \\
DTC~\cite{dtc} &
  (AAAI'21) &
   &
   &
  80.14 &
  67.88 &
  24.08 &
  7.18 \\
MC-Net~\cite{mcnet} &
  (MICCAI'21) &
   &
   &
  80.92 &
  68.90 &
  17.25 &
  2.76 \\
URPC~\cite{urpc} &
  (MedIA'22) &
   &
   &
  80.75 &
  68.54 &
  19.81 &
  4.98 \\
SS-Net~\cite{ssnet} &
  (MICCAI'22) &
   &
   &
  83.33 &
  71.79 &
  15.70 &
  4.33 \\
MC-Net+~\cite{mcnet_plus} &
  (MedIA'22) &
   &
   &
  83.23 &
  71.70 &
  14.92 &
  3.43 \\
BCP~\cite{copy_paste} &
  (CVPR'23) &
   &
   &
  {\color[HTML]{0000FF} 87.52} &
  {\color[HTML]{0000FF} 78.15} &
  \color[HTML]{0000FF} {8.41}&
  2.64 \\
UniMatch~\cite{unimatch} &
  (CVPR'23) &
   &
   &
  86.08 &
  75.83 &
  12.04 &
  2.85 \\
CAML~\cite{CAML} &
  (MICCAI'23) &
   &
   &
  87.34 &
  77.65 &
  {9.76} &
  {\color[HTML]{0000FF} 2.49} \\
Ours &
   &
   &
   &
  {\color[HTML]{FF0000} 89.98} &
  {\color[HTML]{FF0000} 81.84} &
  {\color[HTML]{FF0000} 5.90} &
  {\color[HTML]{FF0000} 1.85} \\ \midrule
UA-MT~\cite{uamt} &
  (MICCAI'19) &
  8(10\%) &
  72(90\%) &
  85.81 &
  75.41 &
  18.25 &
  5.04 \\
SASSNet~\cite{sassnet} &
  (MICCAI'20) &
   &
   &
  85.71 &
  75.35 &
  14.74 &
  4.00 \\
DTC~\cite{dtc} &
  (AAAI'21) &
   &
   &
  84.55 &
  73.91 &
  13.80 &
  3.69 \\
MC-Net~\cite{mcnet} &
  (MICCAI'21) &
   &
   &
  86.87 &
  78.49 &
  11.17 &
  2.18 \\
URPC~\cite{urpc} &
  (MedIA'22) &
   &
   &
  83.37 &
  71.99 &
  17.91 &
  4.41 \\
SS-Net~\cite{ssnet} &
  (MICCAI'22) &
   &
   &
  86.56 &
  76.61 &
  12.76 &
  3.02 \\
MC-Net+~\cite{mcnet_plus} &
  (MedIA'22) &
   &
   &
  87.68 &
  78.27 &
  10.35 &
  1.85 \\
DMD~\cite{dmd} &
  (MICCAI'23) &
   &
   &
  {89.70} &
  {\color[HTML]{0000FF} 81.42} &
  {\color[HTML]{0000FF} 6.88} &
  1.78 \\
BCP~\cite{copy_paste} &
  (CVPR'23) &
   &
   &
  89.55 &
  81.22 &
  7.10 &
  {\color[HTML]{0000FF} 1.69} \\
UniMatch~\cite{unimatch} &
  (CVPR'23) &
   &
   &
  89.09 &
  80.47 &
  12.50 &
  3.59 \\
CAML~\cite{CAML} &
  (MICCAI'23) &
   &
   &
  89.62 &
  81.28 &
  8.76 &
  2.02 \\
RCPS~\cite{rcps_jbhi24} &
  (JBHI'24) &
   &
   &
  \color[HTML]{0000FF}{90.73} &
  - &
  7.91 &
  2.05 \\
Ours &
   &
   &
   &
  {\color[HTML]{FF0000} 91.33} &
  {\color[HTML]{FF0000} 84.11} &
  {\color[HTML]{FF0000} 5.29} &
  {\color[HTML]{FF0000} 1.53} \\ \midrule
UA-MT~\cite{uamt} &
  (MICCAI'19) &
  16(20\%) &
  64(80\%) &
  88.18 &
  79.09 &
  9.66 &
  2.62 \\
SASSNet~\cite{sassnet} &
  (MICCAI'20) &
   &
   &
  88.11 &
  79.08 &
  12.31 &
  3.27 \\
DTC~\cite{dtc} &
  (AAAI'21) &
   &
   &
  87.79 &
  78.52 &
  10.29 &
  2.50 \\
MC-Net~\cite{mcnet} &
  (MICCAI'21) &
   &
   &
  90.43 &
  82.69 &
  6.52 &
  1.66 \\
URPC~\cite{urpc} &
  (MedIA'22) &
   &
   &
  87.68 &
  78.36 &
  14.39 &
  3.52 \\
SS-Net~\cite{ssnet} &
  (MICCAI'22) &
   &
   &
  88.19 &
  79.21 &
  8.12 &
  2.20 \\
MC-Net+~\cite{mcnet_plus} &
  (MedIA'22) &
   &
   &
  90.60 &
  82.93 &
  6.27 &
  {\color[HTML]{0000FF} 1.58} \\
DMD~\cite{dmd} &
  (MICCAI'23) &
   &
   &
  90.46 &
  82.66 &
  6.39 &
  1.62 \\
BCP~\cite{copy_paste} &
  (CVPR'23) &
   &
   &
  90.18 &
  82.36 &
  6.64 &
  1.61 \\
UniMatch~\cite{unimatch} &
  (CVPR'23) &
   &
   &
  90.77 &
  83.18 &
  7.21 &
  2.05 \\
CAML~\cite{CAML} &
  (MICCAI'23) &
   &
   &
  90.78 &
  {\color[HTML]{0000FF} 83.19} &
  {\color[HTML]{0000FF} 6.11} &
  1.68 \\
BSNet~\cite{bsnet_tmi24} &
  (TMI'24) &
   &
   &
  90.43 &
  - &
  6.21 &
  1.63 \\
RCPS~\cite{rcps_jbhi24} &
  (JBHI'24) &
   &
   &
  {\color[HTML]{0000FF} 91.21} &
  - &
  6.54 &
  1.81 \\
Ours &
   &
   &
   &
  {\color[HTML]{FF0000} 91.61} &
  {\color[HTML]{FF0000} 84.57} &
  {\color[HTML]{FF0000} 5.36} &
  {\color[HTML]{FF0000} 1.57} \\ \bottomrule
\end{tabular}
\end{table*}
\begin{table*}[]
\centering\caption{Comparisons on the ACDC dataset. "$\uparrow$" and "$\downarrow$" indicate the larger and the smaller the better, respectively.}\label{tab:acdc_results}
\begin{tabular}{@{}ll|ll|l|l|l|l@{}}
\toprule
\multicolumn{2}{c|}{} &
  \multicolumn{2}{c|}{\#Scans used} &
  \multicolumn{4}{c}{{Metrics}} \\ \cmidrule(l){3-8} 
\multicolumn{2}{c|}{\multirow{-2}{*}{{Method}}} &
  {Lab.} &
  \multicolumn{1}{l|}{{Unlab.}} &
  {Dice(\%)$\uparrow$} &
  {Jaccard(\%)$\uparrow$} &
  {95HD(voxel)$\downarrow$} &
  {ASD(voxel)$\downarrow$ }\\ \midrule
U-Net &
   &
  3(5\%) &
  0 &
  47.83 &
  37.01 &
  31.16 &
  12.62 \\
U-Net &
   &
  7(10\%) &
  0 &
  79.41 &
  68.11 &
  9.35 &
  2.70 \\
U-Net &
   &
  70(All) &
  0 &
  91.44 &
  84.59 &
  4.30 &
  0.99 \\ \midrule
UA-MT~\cite{uamt} &
  (MICCAI'19) &
  3(5\%) &
  67(95\%) &
  46.04 &
  35.97 &
  20.08 &
  7.75 \\
SASSNet~\cite{sassnet} &
  (MICCAI'20) &
   &
   &
  57.77 &
  46.14 &
  20.05 &
  6.06 \\
DTC~\cite{dtc} &
  (AAAI'21) &
   &
   &
  56.90 &
  45.67 &
  23.36 &
  7.39 \\
MC-Net~\cite{mcnet} &
  (MICCAI'21) &
   &
   &
  62.85 &
  52.29 &
  7.62 &
  2.33 \\
URPC~\cite{urpc} &
  (MedIA'22) &
   &
   &
  55.87 &
  44.64 &
  13.60 &
  3.74 \\
SS-Net~\cite{ssnet} &
  (MICCAI'22) &
   &
   &
  65.82 &
  55.38 &
  6.67 &
  2.28 \\
DMD~\cite{dmd} &
  (MICCAI'23) &
   &
   &
  80.60 &
  69.08 &
  5.96 &
  1.90 \\
UniMatch~\cite{unimatch} &
  (CVPR'23) &
   &
   &
  {\color[HTML]{0000FF} 84.38} &
  {\color[HTML]{0000FF} 75.54} &
  {\color[HTML]{0000FF} 5.06} &
  {\color[HTML]{0000FF} 1.04} \\
URCA~\cite{urca_cmpb24} &
  (CMPB'24) &
   &
   &
  83.31   &
  - &
  6.95   &
  2.16   \\
Ours &
   &
   &
   &
  {\color[HTML]{FF0000} 88.27} &
  {\color[HTML]{FF0000} 80.17} &
  {\color[HTML]{FF0000} 1.53} &
  {\color[HTML]{FF0000} 0.46} \\ \midrule
UA-MT~\cite{uamt} &
  (MICCAI'19) &
  7(10\%) &
  63(90\%) &
  81.65 &
  70.64 &
  6.88 &
  2.02 \\
SASSNet~\cite{sassnet} &
  (MICCAI'20) &
   &
   &
  84.50 &
  74.34 &
  5.42 &
  1.86 \\
DTC~\cite{dtc} &
  (AAAI'21) &
   &
   &
  84.29 &
  73.92 &
  12.81 &
  4.01 \\
MC-Net~\cite{mcnet} &
  (MICCAI'21) &
   &
   &
  86.44 &
  77.04 &
  5.50 &
  1.84 \\
URPC~\cite{urpc} &
  (MedIA'22) &
   &
   &
  83.10 &
  72.41 &
  4.84 &
  1.53 \\
SS-Net~\cite{ssnet} &
  (MICCAI'22) &
   &
   &
  86.78 &
  77.67 &
  6.07 &
  1.40 \\
DMD~\cite{dmd} &
  (MICCAI'23) &
   &
   &
  87.52 &
  78.62 &
  4.81 &
  1.60 \\
UniMatch~\cite{unimatch} &
  (CVPR'23) &
   &
   &
  {\color[HTML]{0000FF} 88.08} &
  {\color[HTML]{0000FF} 80.10} &
  {\color[HTML]{0000FF} 2.09} &
  {\color[HTML]{FF0000} 0.45} \\
URCA~\cite{urca_cmpb24} &
  (CMPB'24) &
   &
   &
  87.86  &
  - &
  4.21  &
  1.36  \\
Ours &
   &
   &
   &
  {\color[HTML]{FF0000} 89.08} &
  {\color[HTML]{FF0000} 81.44} &
  {\color[HTML]{FF0000} 1.52} &
  {\color[HTML]{0000FF} 0.52}\\ \bottomrule
\end{tabular}
\end{table*}
\begin{table*}[]
    \centering\caption{{Comparisons on the Pancreas-CT dataset. "$\uparrow$" and "$\downarrow$" indicate the larger and the smaller the better, respectively.}}\label{tab:pct_results}
\begin{tabular}{@{}ll|ll|l|l|l|l@{}}
\toprule
\multicolumn{2}{c|}{} &
  \multicolumn{2}{c|}{\#Scans used} &
  \multicolumn{4}{c}{{Metrics}} \\ \cmidrule(l){3-8} 
\multicolumn{2}{c|}{\multirow{-2}{*}{{Method}}} &
  {Lab.} &
  \multicolumn{1}{l|}{{Unlab.}} &
  {Dice(\%)$\uparrow$} &
  {Jaccard(\%)$\uparrow$} &
  {95HD(voxel)$\downarrow$} &
  {ASD(voxel)$\downarrow$ }\\ \midrule
V-Net &
   &
  6(10\%) &
  0 &
  42.56 &
  36.71 &
  42.61 &
  11.68 \\
V-Net &
   &
  12(20\%) &
  0 &
  62.42 &
  48.06 &
  22.34 &
  4.77 \\
V-Net &
   &
  62(All) &
  0 &
  77.84 &
  64.78 &
  8.92 &
  3.73 \\ \midrule
UA-MT~\cite{uamt} &
  (MICCAI'19) &
  6(10\%) &
  56(90\%) &
  53.07 &
  38.90 &
  25.11 &
  9.22 \\
SASSNet~\cite{sassnet} &
  (MICCAI'20) &
   &
   &
  56.23 &
  41.98 &
  26.16 &
  8.97 \\
DTC~\cite{dtc} &
  (AAAI'21) &
   &
   &
  59.54 &
  45.61 &
  16.53 &
  3.12 \\
UniMatch~\cite{unimatch} &
  (CVPR'23) &
   &
   &
  69.90 &
  {\color[HTML]{0000FF} 55.13} &
  {\color[HTML]{0000FF} 12.94} &
  3.56 \\
RCPS~\cite{rcps_jbhi24} &
  (JBHI'24) &
   &
   &
  {\color[HTML]{0000FF} 76.62} &
  - &
  16.32 &
  {\color[HTML]{0000FF} 3.01} \\
Ours &
   &
   &
   &
  {\color[HTML]{FF0000} 79.69} &
  {\color[HTML]{FF0000} 66.93} &
  {\color[HTML]{FF0000} 11.18} &
  {\color[HTML]{FF0000} 2.64} \\ \midrule
UA-MT~\cite{uamt} &
  (MICCAI'19) &
  12(20\%) &
  50(80\%) &
  72.43 &
  57.91 &
  11.01 &
  4.25 \\
SASSNet~\cite{sassnet} &
  (MICCAI'20) &
   &
   &
  70.47 &
  55.74 &
  10.95 &
  4.26 \\
DTC~\cite{dtc} &
  (AAAI'21) &
   &
   &
  75.94 &
  62.41 &
  {8.25} &
  2.21 \\
UniMatch~\cite{unimatch} &
  (CVPR'23) &
   &
   &
  79.52 &
  {\color[HTML]{0000FF} 66.64} &
  13.05 &
  3.02 \\
RCPS~\cite{rcps_jbhi24} &
  (JBHI'24) &
   &
   &
  {\color[HTML]{0000FF} 81.59} &
  {-} &
  {\color[HTML]{0000FF} 7.50} &
  {\color[HTML]{0000FF} 2.03} \\
Ours &
   &
   &
   &
  {\color[HTML]{FF0000} 83.13} &
  {\color[HTML]{FF0000} 71.46} &
  {\color[HTML]{FF0000} 5.20} &
  {\color[HTML]{FF0000} 1.88} \\ \bottomrule
\end{tabular}
\end{table*}
\begin{table*}[]
\centering\caption{{Comparisons on the ISIC 2018 dataset. "$\uparrow$" and "$\downarrow$" indicate the larger and the smaller the better, respectively.}}\label{tab:isic2018_results}
\begin{tabular}{@{}ll|ll|l|l|l|l@{}}
\toprule
\multicolumn{2}{c|}{} &
  \multicolumn{2}{c|}{\#Scans used} &
  \multicolumn{4}{c}{{Metrics}} \\ \cmidrule(l){3-8} 
\multicolumn{2}{c|}{\multirow{-2}{*}{{Method}}} &
  {Lab.} &
  \multicolumn{1}{l|}{{Unlab.}} &
  {Dice(\%)$\uparrow$} &
  {Jaccard(\%)$\uparrow$} &
  {95HD(voxel)$\downarrow$} &
  {ASD(voxel)$\downarrow$ }\\ \midrule
U-Net &
   &
  129(5\%) &
  0 &
  60.75 &
  49.41 &
  82.28 &
  38.69 \\
U-Net &
   &
  259(10\%) &
  0 &
  76.16 &
  66.55 &
  43.76 &
  17.21 \\
U-Net &
   &
  518(20\%) &
  0 &
  81.18 &
  71.54 &
  35.73 &
  14.22 \\
U-Net &
   &
  2594(All) &
  0 &
  86.84 &
  78.81 &
  24.58 &
  9.90 \\ \midrule
UA-MT~\cite{uamt} &
  (MICCAI'19) &
  129(5\%) &
  2465(95\%) &
  70.22 &
  58.77 &
  59.91 &
  24.96 \\
FixMatch~\cite{fixmatch} &
  (NeurIPS'20) &
   &
   &
  83.15 &
  73.54 &
  30.40 &
  {{\color[HTML]{0000FF} 12.02}} \\
DTC~\cite{dtc} &
  (AAAI'21) &
   &
   &
  76.59 &
  66.13 &
  49.52 &
  20.65 \\
UniMatch~\cite{unimatch} &
  (CVPR'23) &
   &
   &
  {{\color[HTML]{0000FF} 83.44}} &
  {{\color[HTML]{0000FF} 74.15}} &
  {{\color[HTML]{0000FF} 30.37}} &
  12.33 \\
Ours &
   &
   &
   &
  {\color[HTML]{FF0000} 84.10} &
  {\color[HTML]{FF0000} 74.69} &
  {\color[HTML]{FF0000} 28.60} &
  {\color[HTML]{FF0000} 11.28} \\ \midrule
UA-MT~\cite{uamt} &
  (MICCAI'19) &
  259(10\%) &
  2335(90\%) &
  77.88 &
  68.76 &
  46.15 &
  17.80 \\
FixMatch~\cite{fixmatch} &
  (NeurIPS'20) &
   &
   &
  83.42 &
  73.77 &
  27.28 &
  10.98 \\
DTC~\cite{dtc} &
  (AAAI'21) &
   &
   &
  79.29 &
  69.36 &
  44.49 &
  18.46 \\
UniMatch~\cite{unimatch} &
  (CVPR'23) &
   &
   &
  {{\color[HTML]{0000FF} 84.19}} &
  {{\color[HTML]{0000FF} 75.19}} &
  {{\color[HTML]{0000FF} 25.57}} &
  {{\color[HTML]{0000FF} 10.08}} \\
Ours &
   &
   &
   &
  {\color[HTML]{FF0000} 84.71} &
  {\color[HTML]{FF0000} 75.81} &
  {\color[HTML]{FF0000} 25.56} &
  {\color[HTML]{FF0000} 9.91} \\ \midrule
UA-MT~\cite{uamt} &
  (MICCAI'19) &
  518(20\%) &
  2076(80\%) &
  82.21 &
  72.47 &
  39.46 &
  16.09 \\
FixMatch~\cite{fixmatch} &
  (NeurIPS'20) &
   &
   &
  84.01 &
 \color[HTML]{0000FF} {75.67} &
  26.98 &
  {{\color[HTML]{0000FF} 9.99}} \\
DTC~\cite{dtc} &
  (AAAI'21) &
   &
   &
  83.07 &
  73.86 &
  31.49 &
  12.37 \\
UniMatch~\cite{unimatch} &
  (CVPR'23) &
   &
   &
  {{\color[HTML]{0000FF} 84.76}} &
  {{75.60}} &
  {{\color[HTML]{0000FF} 25.44}} &
  10.24 \\
Ours &
   &
   &
   &
  {\color[HTML]{FF0000} 85.43} &
  {\color[HTML]{FF0000} 76.68} &
  {\color[HTML]{FF0000} 23.77} &
  {\color[HTML]{FF0000} 9.34} \\ \bottomrule
\end{tabular}
\end{table*}

\begin{figure*}
    \centering\includegraphics[width=0.9\textwidth]{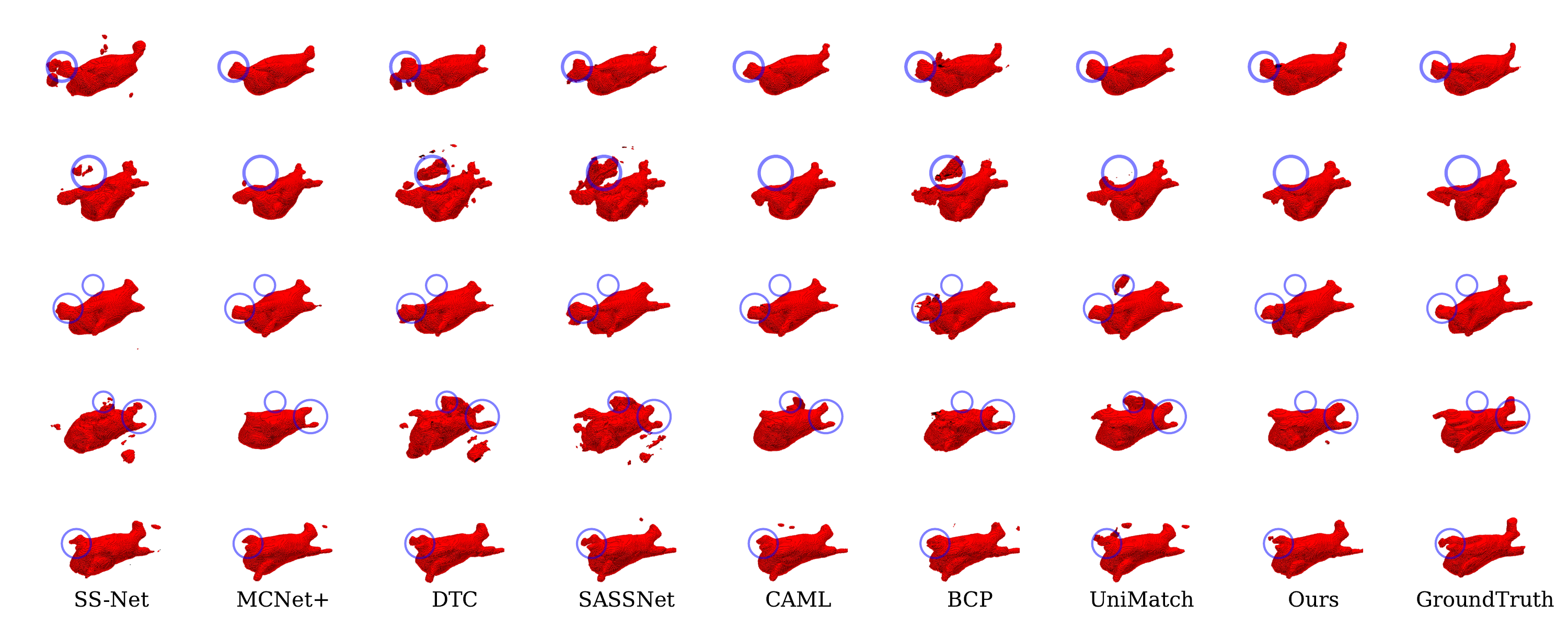}
    \caption{some visualization examples of several semi-supervised segmentation methods with 10\% labeled data and ground truth on the LA dataset.}
    \label{fig:la_3d_result_img}
\end{figure*}

\subsection{Qualitative Comparison}
Fig.~\ref{fig:la_3d_result_img} presents some 3D visualization examples of several compared methods and the corresponding ground truth on the LA dataset. It can be observed that our CrossMatch outperforms other methods in terms of segmentation results. Particularly, our segmentation edges are smoother, with fewer misclassified voxels, and  closer to the ground truth.

{In Fig.~\ref{fig:pct_3d_result}, we visualize the results on the Pancreas-CT dataset. CrossMatch has the smallest gap compared to the ground truth, making its segmentation results more accurate. There are fewer misclassifications and errors, and the segmentation is more continuous.}

\subsection{Quantitative Comparison}
Table \ref{tab:la_results} summarizes the quantitative results and reveals that CrossMatch surpasses state-of-the-art (SOTA) techniques on the LA dataset. When using 5\% of the data with labels (4-label setting), although our Dice and Jaccard are close to those of SOTA methods, CrossMatch achieves significantly better results in the remaining evaluation metrics.
Furthermore, significant performance improvements are realized in the scenarios with 8 and 16 labels. Especially using only 10\% of the data with labels, CrossMatch exceeds the segmentation results obtained by fully supervised learning of V-Net on 100\% of the data with labels, achieving a Dice of 91.33\%.

Quantitative results on the ACDC dataset summarized in Table \ref{tab:acdc_results} further {demonstrate} the effectiveness of CrossMatch.  Particularly our method is more outstanding in terms of performance enhancement, where Dice is increased by 3.89\% in the setting of 3-label. The experimental setups listed in both Table \ref{tab:la_results} and Table \ref{tab:acdc_results} are the same as those in \cite{CAML}, meaning all results are derived from the final iteration outcomes.

{The quantitative results of the Pancreas-CT dataset are shown in Table \ref{tab:pct_results}. 
The results indicate that our CrossMatch outperforms the SOTA methods across all evaluation metrics in every data split. Notably, when using only 10\% of labeled data, the Dice is increased by 3.07\% compared to the SOTA method. When using 20\% of labeled data, the 95HD is improved by 2.3\%.}

{Table~\ref{tab:isic2018_results} shows the results on the ISIC 2018 dataset. Our CrossMatch outperforms all other methods across all data splits and metrics. 
Notably, 
with 20\% labeled data, CrossMatch surpasses the distance metric performance achieved by the fully supervised method.}

\subsection{Computational Performance Analysis}
\begin{table}[H]
\centering\caption{Comparison of iteration times for different methods. Time is recorded from the beginning of data migration to the CUDA device to the end of backpropagation. The time averages are taken after $1k$ iterations.}\label{tab:compute_time}
\begin{tabular}{@{}c|lcc@{}}
\toprule
Time~(ms)$\downarrow$                   & Method  & \#Params~(M)$\downarrow$ & \#Flops~(G)$\downarrow$ \\ \midrule
22                         & V-Net   & 9.443       & 187.409    \\ \midrule
273                        & UA-MT   & 9.443       & 187.409    \\
501                        & SASSNet & 20.463      & 249.194    \\
545                        & DTC     & 9.443       & 187.538    \\
486                        & MC-Net  & 12.348      & 380.394    \\
1269                       & MC-Net+ & 15.247      & 572.229    \\
1057                       & CAML    & 19.725      & 450.677    \\
379                        & BCP     & 9.443       & 187.409    \\
{\color[HTML]{FF0000} 210} & Ours    & 9.443       & 187.409    \\ \bottomrule
\end{tabular}
\end{table}
As described in section~\ref{sec:efficient_compute}, our method also exhibits excellent computational efficiency. For comparative analysis, we have compiled computation performance data from a range of similar works, evaluating them based on their publicly available source codes. The experimental setup is standardized, with all hyperparameters and optimizer configurations identical, and we record the average duration of a single iteration from full data loading onto CUDA devices to the completion of backpropagation, as shown in Table \ref{tab:compute_time}.

After 1000 iterations, as a straightforward, fully supervised learning method,  V-Net requires 22 ms per iteration, whereas these semi-supervised learning methods that require a combination of labeled and unlabeled data for training need more time, such as UA-MT~\cite{uamt},  MC-Net~\cite{mcnet}, MC-Net+~\cite{mcnet_plus} and CAML~\cite{CAML}  take 273 ms, 486 ms, 1269 ms and 1057 ms per iteration, respectively.
In contrast, thanks to its streamlined structure and self-training pipeline, our CrossMatch only requires 210 ms per iteration, which is significantly lower than other semi-supervised segmentation methods, thus highlighting its efficient computational characteristics.

\subsection{Ablation Study}\label{sec:ablation}

\subsubsection{Ablation Study on $\mathcal L_{dkd}$ Components}

\begin{table}[]
\centering\caption{{Ablation study on $\mathcal L^w_{dkd}$ and $\mathcal L^s_{dkd}$. Comparison on the LA dataset.}}\label{tab:abl_dkd}
\resizebox{\linewidth}{!}{%
\begin{tabular}{@{}cc|c|l|l|l|l@{}}
\toprule
\multicolumn{2}{c|}{Components} & \multicolumn{1}{c|}{\#Labeled}    & \multicolumn{4}{c}{Metrics}                         \\ \midrule
$\mathcal L^w_{dkd}$      & $\mathcal L^s_{dkd}$     & \multicolumn{1}{l|}{\#Scans used} &
  Dice(\%)$\uparrow$ &
  Jaccard(\%)$\uparrow$ &
  95HD(voxel)$\downarrow$ &
  ASD(voxel)$\downarrow$ \\ \midrule
  &   &  & 89.26 & 80.71 & 7.35  & 2.17                        \\
\checkmark &   &  & 89.28 & 80.72 & 7.45  & 2.15                        \\
  & \checkmark &  & 89.80 & 81.58 & 5.94  & {\color[HTML]{FF0000} 1.77} \\
\checkmark &
  \checkmark &
  \multirow{-4}{*}{4(5\%)} &
  {\color[HTML]{FF0000} 89.98} &
  {\color[HTML]{FF0000} 81.84} &
  {\color[HTML]{FF0000} 5.90} &
  1.85 \\ \midrule
  &   &  & 86.18 & 76.28 & 10.35 & 2.02                        \\
\checkmark &   &  & 85.66 & 76.51 & 9.32  & 1.94                        \\
  & \checkmark &  & 89.82 & 81.64 & 6.34  & 1.77                        \\
\checkmark &
  \checkmark &
  \multirow{-4}{*}{8(10\%)} &
  {\color[HTML]{FF0000} 91.33} &
  {\color[HTML]{FF0000} 84.11} &
  {\color[HTML]{FF0000} 5.29} &
  {\color[HTML]{FF0000} 1.53} \\ \midrule
  &   &  & 90.98 & 83.53 & 5.78  & 1.85                        \\ 
\checkmark &   &  & 91.46 & 84.33 & 5.50  & 1.63                        \\
  & \checkmark &  & 91.25 & 83.98 & 6.39  & 2.07                        \\
\checkmark &
  \checkmark &
  \multirow{-4}{*}{16(20\%)} &
  {\color[HTML]{FF0000} 91.61} &
  {\color[HTML]{FF0000} 84.57} &
  {\color[HTML]{FF0000} 5.36} &
  {\color[HTML]{FF0000} 1.57} \\ \bottomrule
\end{tabular}%
}
\end{table}
{Tab.~\ref{tab:abl_dkd} presents an ablation study to evaluate the impact of the loss components $\mathcal L^w_{dkd}$ and $\mathcal L^s_{dkd}$ on the performance of our model using the LA dataset. We choose the optimal parameters, and the detailed process can be found in the discussion section. To ensure the results are universal, we conduct experiments across all data splits.}

{When  using 5\%, 10\% and 20\% of labeled data, the addition of $\mathcal L^w_{dkd}$ and $\mathcal L^s_{dkd}$ improves Dice performance by 0.71\%, 5.14\%, and 0.63\%, respectively. Particularly in the cases of 5\% and 10\%, the introduction of $\mathcal L^s_{dkd}$ significantly enhances the model's distance metric performance due to the regularization effect of the large amount of unlabeled data. From the experimental results, it's evident that with the combined effects of $\mathcal L^w_{dkd}$ and $\mathcal L^s_{dkd}$, our segmentation model achieves a dual benefit, leading to relatively precise segmentation.}

\subsubsection{Ablation Study on $\mathcal L_{tkd}$ Components}

{The $\mathcal L_{tkd}$ in Eq.~\ref{eq:tkd} is actually the average of four feedforward streams under the supervision of $p^w_n$. To study the impact of losses from each feedforward stream on performance, we conduct ablation experiments. Specifically, we first remove the supervision of $p^s_w$ and $p^w_s$, as well as $p^w_w$ and $p^s_s$, to verify the necessity of the Cross operation. Then, we sequentially remove the supervision of each of the four feedforward streams, keeping only the remaining three. It's important to note that we didn't conduct experiments with only one supervision term remaining, as this would degrade the method, make the disturbance space too singular and revert it to the original FixMatch.}

\begin{table}[]
\centering\caption{{Impact of $\mathcal L_{tkd}$ supervision components on performance on the LA dataset. $p^i_j$ checkmark indicates the application of the $p^w_n$ supervision loss component. 10\% of labeled data are used for training.}}\label{tab:ssl_loss_sup}
\resizebox{\linewidth}{!}{%
\begin{tabular}{@{}cccc|l|l|l|l@{}}
\toprule
\multicolumn{4}{c|}{Components} & \multicolumn{4}{c}{Metrics}                           \\ \midrule
$p^w_w$     & $p^s_w$    & $p^w_s$    & $p^s_s$    & Dice(\%)$\uparrow$ & Jaccard(\%)$\uparrow$ & 95HD(voxel)$\downarrow$ & ASD(voxel)$\downarrow$ \\\midrule
\checkmark      &       &       & \checkmark     & 90.24     & 82.36        & 8.43         & 2.10        \\ 
       & \checkmark     & \checkmark     &       & 90.32     & 82.45        & 6.59         & 1.81        \\
       & \checkmark     & \checkmark     & \checkmark     & 90.94     & 83.44        & 7.59         & 1.93        \\
\checkmark      &       & \checkmark     & \checkmark     & 89.04     & 80.53        & 7.48         & 1.94        \\
\checkmark      & \checkmark     &       & \checkmark     & 88.55     & 79.65        & 8.15         & 1.93        \\
\checkmark & \checkmark & \checkmark &   & {\color[HTML]{0000FF} 91.03} & {\color[HTML]{0000FF} 83.59} & {\color[HTML]{0000FF} 5.57} & {\color[HTML]{0000FF} 1.62} \\
\checkmark & \checkmark & \checkmark & \checkmark & {\color[HTML]{FF0000} 91.33} & {\color[HTML]{FF0000} 84.11} & {\color[HTML]{FF0000} 5.29} & {\color[HTML]{FF0000} 1.53} \\ \bottomrule
\end{tabular}}
\end{table}

{Results of the melting research are shown in Tab.~\ref{tab:ssl_loss_sup}. The model performs best when all loss terms are used. Specifically, $p^s_w$ and $p^w_s$ can better capture edge details under moderate disturbance intensity, significantly improving distance metrics. Our Cross operation (using $p^s_w, p^w_s$) is notably better than the Dual-Stream Perturbations in~\cite{unimatch} (using $p^w_w, p^s_s$). These results validate the effectiveness of our method.}

\subsubsection{Ablation Study on $\mathcal L_{ip}$ Components}

\begin{table}[]
\centering\caption{{Impact of $\mathcal L_{ip}$ supervision components on performance on the LA dataset. $p^{s_i}_n$ checkmark indicates the application of the $p^w_n$ supervision loss component.}}\label{tab:ip_loss_sup_strong12}
\resizebox{\linewidth}{!}{%
\begin{tabular}{@{}cc|c|l|l|l|l@{}}
\toprule
\multicolumn{2}{c|}{Components} & \multicolumn{1}{l|}{\#Labeled}    & \multicolumn{4}{c}{Metrics}                         \\ \midrule
$p^{s_1}_n$             & $p^{s_2}_n$            & \multicolumn{1}{l|}{\#Scans used} & Dice(\%)↑ & Jaccard(\%)↑ & 95HD(voxel)↓ & ASD(voxel)↓ \\ \midrule
               &               &                                 & 88.44    & 79.38       & 7.61         & 2.24        \\
\checkmark              &               &                                 & 88.71    & 79.80       & 7.04         & 2.15        \\
               & \checkmark             &                                 & 89.25    & 80.65       & 7.77         & 2.58        \\
\checkmark & \checkmark & \multirow{-4}{*}{4(5\%)}   & {\color[HTML]{FF0000} 89.98} & {\color[HTML]{FF0000} 81.84} & {\color[HTML]{FF0000} 5.90} & {\color[HTML]{FF0000} 1.85} \\ \midrule
               &               &                                 & 86.73    & 77.20       & 8.99         & 2.11        \\
\checkmark              &               &                                 & 90.38    & 82.53       & 6.07         & 1.70        \\
               & \checkmark             &                                 & 89.47    & 81.17       & 7.23         & 1.91        \\
\checkmark & \checkmark & \multirow{-4}{*}{8(10\%)}  & {\color[HTML]{FF0000} 91.33} & {\color[HTML]{FF0000} 84.11} & {\color[HTML]{FF0000} 5.29} & {\color[HTML]{FF0000} 1.53} \\\midrule
               &               &                                 & 90.87    & 83.34       & 5.87         & 1.85        \\
\checkmark              &               &                                 & 90.74    & 83.13       & 5.89         & 1.68        \\
               & \checkmark             &                                 & 90.84    & 83.31       & 5.80         & 1.70        \\
\checkmark & \checkmark & \multirow{-4}{*}{16(20\%)} & {\color[HTML]{FF0000} 91.61} & {\color[HTML]{FF0000} 84.57} & {\color[HTML]{FF0000} 5.36} & {\color[HTML]{FF0000} 1.57} \\ \bottomrule
\end{tabular}%
}
\end{table}

{As shown in Fig.~\ref{fig:overview}, our overall loss also includes pseudo-label supervision for two strongly perturbed feedforward streams. To explore the impact of these two supervisions on performance, we conduct an ablation study, and the results are shown in Tab.~\ref{tab:ip_loss_sup_strong12}.}

{The results show that the model performs best when applying all strong perturbation feeds, indicating that incorporating multiple strong perturbation feeds is necessary.}

\subsection{Discussion} \label{sec:discussion}

\subsubsection{Impact of Supervised Loss Type}

\begin{table}[]
\centering\caption{{The impact of $\mathcal L_{sup}$ loss types on performance on the LA dataset. 10\% of labeled data are used for training. Mix represents the average of CE and Dice losses.}}\label{tab:sup_loss_type}
\begin{tabular}{@{}c|c|c|c|c@{}} \toprule
\multicolumn{1}{c|}{} & \multicolumn{4}{c}{Metrics} \\ \cmidrule(l){2-5}
\multicolumn{1}{c|}{\multirow{-2}{*}{Loss Type}} & Dice(\%)↑                    & Jaccard(\%)↑                 & 95HD(voxel)↓                & ASD(voxel)↓                 \\ \midrule
CE                   & 91.01 & 83.57 & 5.47 & 1.81 \\
Dice                 & 91.08 & 83.68 & 5.37 & 1.65 \\
Mix                                        & {\color[HTML]{FF0000} 91.33} & {\color[HTML]{FF0000} 84.11} & {\color[HTML]{FF0000} 5.29} & {\color[HTML]{FF0000} 1.53} \\ \bottomrule
\end{tabular}
\end{table}

{By default, the fully-supervised loss is the average of cross-entropy and Dice loss. To study the impact of the types of fully-supervised loss, we examine the cases of using only cross-entropy and only Dice loss, as shown in Tab.~\ref{tab:sup_loss_type}. The results indicate that the model performance is best with the mixed loss. Specifically, cross-entropy loss performs poorly when dealing with imbalanced classes; Dice loss is better at handling small targets but lacks in global optimization. Thus, the mixed loss that combines the strengths of both shows greater robustness and stability in various scenarios.}

\subsubsection{Impact of Decoder Distillation Loss Type}
\begin{table}[]
\centering\caption{{The impact of $\mathcal L_{dkd}$ loss types on performance on the LA dataset.}}\label{tab:loss_type_abl}
\resizebox{\linewidth}{!}{%
\begin{tabular}{@{}l|c|l|l|l|l@{}} \toprule
\multicolumn{1}{c|}{} &
  \multicolumn{1}{l|}{\#Labeled} &
  \multicolumn{4}{c}{Metrics} \\ \cmidrule(l){2-6}
\multicolumn{1}{c|}{\multirow{-2}{*}{\makecell{Loss Type}}} &
  \multicolumn{1}{c|}{\#Scans used} &
  Dice(\%)$\uparrow$ &
  Jaccard(\%)$\uparrow$ &
  95HD(voxel)$\downarrow$ &
  ASD(voxel)$\downarrow$ \\ \midrule
KL($T=1$) &  & 89.59 & 81.20 & 6.87                        & 2.08 \\
KL($T=2$) &  & 89.77 & 81.49 & {\color[HTML]{FF0000} 5.72} & 1.90 \\
CE      &  & 88.55 & 79.58 & 7.71                        & 1.99 \\
Dice &
  \multirow{-4}{*}{4(5\%)} &
  {\color[HTML]{FF0000} 89.98} &
  {\color[HTML]{FF0000} 81.84} &
  5.90 &
  {\color[HTML]{FF0000} 1.85} \\ \midrule
KL($T=1$) &  & 89.36 & 80.99 & 7.26                        & 1.97 \\
KL($T=2$) &  & 88.99 & 80.57 & 7.01                        & 1.98 \\
CE      &  & 89.27 & 80.85 & 7.12                        & 1.88 \\
Dice &
  \multirow{-4}{*}{8(10\%)} &
  {\color[HTML]{FF0000} 91.33} &
  {\color[HTML]{FF0000} 84.11} &
  {\color[HTML]{FF0000} 5.29} &
  {\color[HTML]{FF0000} 1.53} \\\midrule
KL($T=1$) &  & 90.74 & 83.13 & 5.99                        & 1.70 \\
KL($T=2$) &  & 90.82 & 83.26 & 5.95                        & 1.73 \\
CE      &  & 91.03 & 83.62 & 5.78                        & 1.84 \\
Dice &
  \multirow{-4}{*}{16(20\%)} &
  {\color[HTML]{FF0000} 91.61} &
  {\color[HTML]{FF0000} 84.57} &
  {\color[HTML]{FF0000} 5.36} &
  {\color[HTML]{FF0000} 1.57}\\ \bottomrule
\end{tabular} %
}
\end{table}

{In knowledge distillation, commonly used loss functions include KL divergence and cross-entropy loss. According to research by ~\cite{dmd}, Dice loss also performs well in the field of medical image segmentation. To explore the impact of different loss functions on segmentation performance, we choose the aforementioned three for our experiments. Studies such as ~\cite{liang2024neighbor, self_kd} indicate that the performance of KL divergence is affected by the temperature coefficient $T$, defined as $\sum_j(\mathrm{KL}(p^{w}_{j} / T \parallel  p^{s}_{j} / T)/2 + \mathrm{KL}(p^{s}_{j} / T \parallel  p^{w}_{j} / T)/2)$, where $j\in\{w,s\}$. We set $T=1$ and $T=2$ in our experiments.}

{We conduct experiments on the LA dataset, with results shown in Tab.~\ref{tab:loss_type_abl}. The results indicate that the model performs best when using Dice loss, while KL divergence's performance is unstable at different $T$ values, and the cross-entropy loss performs poorly. Overall, Dice loss is more suitable for medical image segmentation. 
Cross-entropy loss, due to its inadequate detail capturing ability, also fails to stand out in segmentation tasks.}

\subsubsection{Exploring Dropout Types}
\begin{figure}
    \centering
    \includegraphics[width=\linewidth]{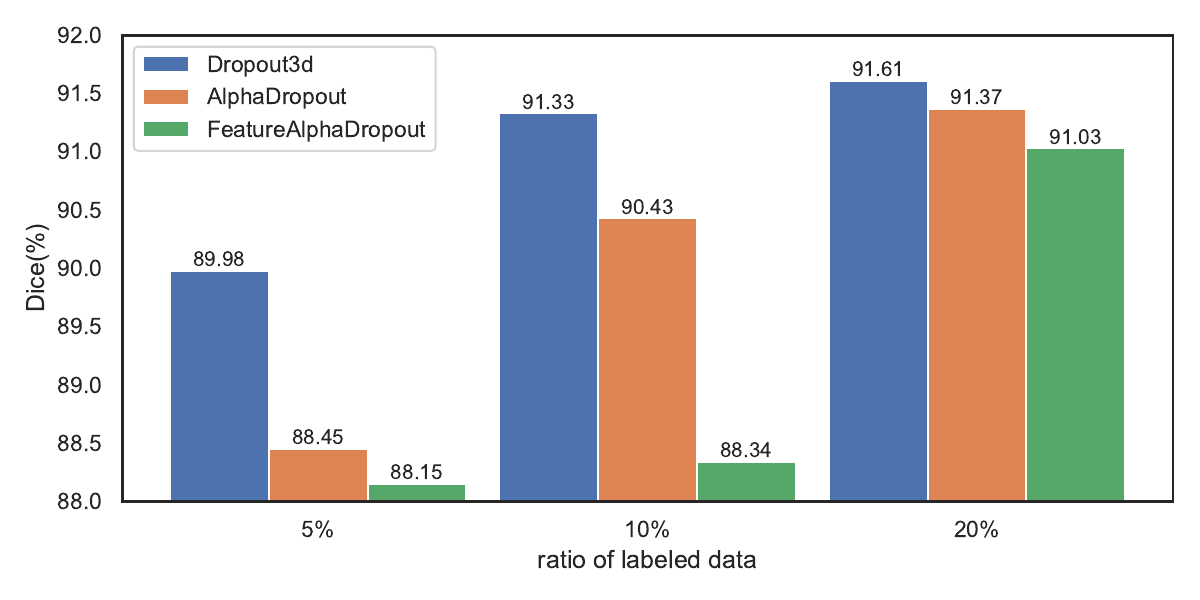}
    \caption{The efficacy of various feature perturbation strategies in our method.}
    \label{fig:abl_drop_type}
\end{figure}

Fig.~\ref{fig:abl_drop_type} shows the results on the type of Dropout used under different training sample ratios on the LA dataset. We investigate three different types of Dropout available in PyTorch: standard Dropout3D, AlphaDropout and FeatureAlphaDropout \cite{alpha_drop}.
It is evident that using the standard Dropout3D as our feature perturbation strategy results in the best performance across all three data splits, followed by AlphaDropout and FeatureAlphaDropout. This may be due to the latter two inducing stronger feature perturbations, resulting in an increased Performance Gap between decoders, which is detrimental to the correct transfer of knowledge in knowledge distillation.

\subsubsection{Effects of Hyperparameter $\eta$}
\begin{table}[htbp]
\centering\caption{The effects of $\eta$ at all metrics on the LA dataset. 10\% of labeled data are used for training.}\label{tab:eta_abl}
\begin{tabular}{@{}c|l|l|l|l@{}}
\toprule
\quad $\eta$ & Dice(\%)$\uparrow$                      & Jaccard(\%)$\uparrow$                   & 95HD(voxel)$\downarrow$                & ASD(voxel)$\downarrow$                 \\ \midrule
\quad 0.10  & 90.88 & 83.35 & 7.80 & 1.93 \\
\quad 0.15 & 90.55 & 82.92 & 7.93 & 2.51 \\
\quad 0.20  & 91.03 & 83.62 & 6.44 & 1.99 \\
\quad 0.25 & 91.28 & 84.01 & 5.77 & 1.62 \\
\quad 0.30 & {\color[HTML]{FF0000} 91.33} & {\color[HTML]{FF0000} 84.11} & {\color[HTML]{FF0000} 5.29} & {\color[HTML]{FF0000} 1.53} \\
\quad 0.35 & 90.78 & 83.21 & 6.91 & 1.88 \\
\quad 0.40  & 87.25 & 77.80 & 9.80 & 2.37 \\
\quad 0.45 & 90.53 & 82.78 & 6.54 & 1.83 \\
\quad 0.50  & 87.78 & 79.22 & 6.76 & 1.90 \\ \bottomrule
\end{tabular}
\end{table}
Tab.~\ref{tab:eta_abl} displays the results of experiments on the parameter of $\eta$ in the setting of  10\% of the data with labels on the LA dataset. The results indicate that the $\eta$ value of 0.3 yields the best performance, surpassing other values across all metrics. Consequently, we have set $\eta$ at 0.3 for all experiments in this study.

\subsubsection{Effects of Hyperparameter $\tau$}
\begin{figure}
    \centering
    \includegraphics[width=\linewidth]{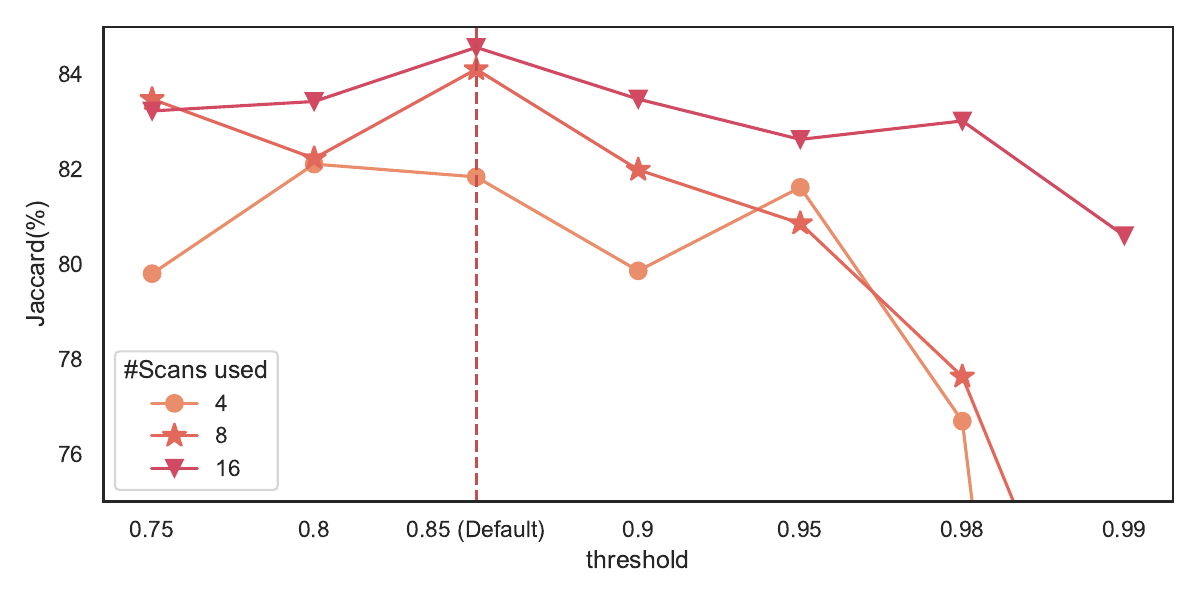}
    \caption{The effect of hyperparameter $\tau$ on performance on the LA dataset.}
    \label{fig:abl_tau}
\end{figure}

{As shown in Fig.~\ref{fig:abl_tau},  the performance of our method on the LA dataset fluctuates with changes in the hyperparameter $\tau$. When the $\tau$ value is low, the model performs poorly, likely because it can't effectively distinguish between correct and incorrect knowledge. As the $\tau$ value increases, the model's performance improves, indicating that a higher $\tau$ can mitigate the transfer of incorrect knowledge. However, as $\tau$ exceeds a certain threshold, performance starts to decline, suggesting that an excessively high $\tau$ prevents the model from effectively learning from the teacher model's knowledge. 
Hence we fix $\tau = 0.85$ in CrossMatch.}

\subsubsection{Impact of Decoder Performance Gap}

\begin{table}[htbp]
\centering\caption{The performance gap between strong and weak decoders on the LA dataset. 10\% of labeled data are used for training.}\label{tab:gap}
\resizebox{\linewidth}{!}{%
\begin{tabular}{@{}ccc|l|l|l|l@{}}\toprule
Bottom & Top   & \makecell[c]{Perf. Gap} & Dice(\%)$\uparrow$ & Jaccard(\%)$\uparrow$ & 95HD(voxel)$\downarrow$ & ASD(voxel)$\downarrow$                 \\ \midrule
0.500 & 0.500 & 0.00 & {\color[HTML]{0070C0} 91.20} & {\color[HTML]{0070C0} 83.87} & {\color[HTML]{0070C0} 5.41} & 1.80                        \\
0.375  & 0.625 & 0.25            & 91.07     & 83.65        & 5.47         & {\color[HTML]{0070C0} 1.59} \\
0.250 & 0.750 & 0.50 & {\color[HTML]{FF0000} 91.33} & {\color[HTML]{FF0000} 84.11} & {\color[HTML]{FF0000} 5.29} & {\color[HTML]{FF0000} 1.53} \\
0.125  & 0.875 & 0.75            & 90.22     & 82.32        & 6.59         & 1.76                       \\ \bottomrule
\end{tabular}%
}
\end{table}
Tab.~\ref{tab:gap} presents the results of the performance gap between decoders using 10\% of the data with labels on the LA dataset. It is observed that the optimal model performance is achieved when the performance gap is 0.5. Notably, we also explore the scenario where the decoders are completely consistent, that is when the performance gap is zero. In this case, the strong perturbation decoder and the weak perturbation decoder apply identical perturbations, and the model degenerates into a UniMatch with an additional $\mathcal{L}_{dkd}$.

\subsubsection{{Wrongly Segmented Examples Analysis}}
\begin{figure}
    \centering
    \includegraphics[width=\linewidth]{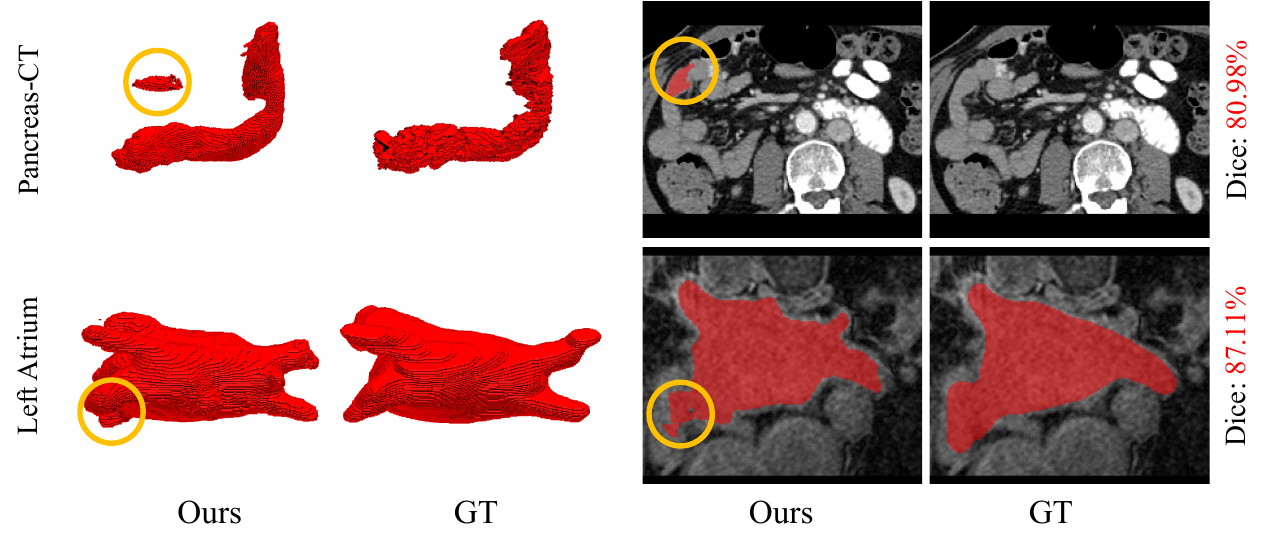}
    \caption{Some wrongly segmented examples using 10\% labeled data on the Pancreas-CT dataset and LA dataset, respectively.}
    \label{fig:wrong_cases}
\end{figure}

{In this subsection, we examine two instances where our segmentation model demonstrates suboptimal performance, as reflected by lower Dice. These instances involve the segmentation of the Pancreas-CT dataset and the LA dataset. Through a detailed analysis of these cases, we aim to uncover the underlying causes of the segmentation errors and propose viable solutions to improve the accuracy of our model.}

{In the first Pancreas-CT case, the model's Dice is 80.98\%. The main challenge is insufficient segmentation of 
the} {pan}{creatic tail.} {
The similarity in intensity between the pancreatic tail and adjacent tissues makes it hard to distinguish, and the small and variable structure of the pancreatic tail causes the model's performance to be unstable in that area. In the second LA case, 
the model's Dice is 87.11\%. The main issue is inaccurate boundary definition caused by partial volume effects near the atrial appendages. Partial volume effects are a common problem in MRI imaging, where the signal from multiple tissue types within a single voxel gets averaged, blurring the boundaries. The anatomical complexity of the left atrium and its appendages also increases the difficulty of segmentation. Additionally, the feature-level and image-level perturbations used in our pseudo-label learning may have further interfered with the model's ability to learn about the blurred boundaries.}

{To address these issues, we propose two solutions. First, advanced image processing techniques, such as multi-scale analysis or deep learning architectures that capture fine details can improve feature extraction capabilities and segmentation accuracy. However, for a fair comparison, we did not make any changes to the original V-Net model in this paper. Second, integrating advanced modelling techniques like shape constraint models or active contour models can enhance the model's accuracy in segmenting complex structures. 
We are committed to continuing to optimize our method to ensure its effectiveness and reliability in clinical applications in future work. }

\subsubsection{Advantages and Limitations}
{In this study, our proposed CrossMatch shows significant performance improvement in various medical image segmentation tasks. The key advantage lies in the introduction of self-knowledge distillation and multiple perturbation strategies, which make full use of limited labeled data and a large amount of unlabeled data, greatly enhancing the model's robustness and accuracy. Experimental results demonstrate that CrossMatch outperforms existing state-of-the-art methods on multiple benchmark datasets, particularly excelling in edge accuracy and generalization ability.}

{However, CrossMatch still has its limitations. Firstly, although the implementation is efficient, the computational resources required for processing large-scale 3D medical images remain high, such as the computation of multiple forward passes. Secondly, its performance still falls short in some complex structure segmentation tasks, especially in areas with high variability and fuzzy boundaries.
Future research could consider incorporating more advanced image processing techniques and modelling methods to further improve the model's accuracy and robustness.}

\section{Conclusion}
We have re-evaluated the role of Self-Knowledge in semi-supervised medical image segmentation and cleverly integrated feature perturbation, consistency regularization and Knowledge Distillation to propose a Self-Training segmentation method named CrossMatch.  We rethink the role of perturbations in semi-supervised tasks and suggest using multiple equivalent encoders and decoders to play roles at different learning stages to expand the traditional teacher-student model, aiming to reduce the capability gap between different roles. Specifically, we derive two encoders from image-level perturbations and three decoders from feature-level perturbations, designating the unperturbed feed-forward flow as the teacher to perform knowledge distillation on the four groups of outcomes produced by the aforementioned encoder and decoder combinations. Additionally, we utilize the properties of Mini Batches to optimize the performance of our method and provide a quantitative iteration time comparison table. Our CrossMatch demonstrates robust performance on four benchmark datasets (LA, ACDC, Pancreas-CT and ISIC-2018), showing significant improvements over SOTA methods. Extensive ablation studies further validate the assumptions and design of our method.

\section*{References}
\bibliographystyle{IEEEtran}

\end{document}